\documentclass[5p,twocolumn]{elsarticle}

\usepackage[utf8]{inputenc}
\usepackage{amsmath,amssymb,mathtools}
\usepackage{paralist}
\usepackage[detect-weight=true, binary-units=true]{siunitx}
\usepackage{url,hyperref,cleveref}
\usepackage[linesnumbered]{algorithm2e}
\crefname{algocf}{algorithm}{algorithms}
\Crefname{algocf}{Algorithm}{Algorithms}
\usepackage{color,colortbl,etoolbox,xcolor} 
\robustify\bfseries
\usepackage{multirow,booktabs}
\usepackage[caption=false]{subfig}
\usepackage{csquotes} 
\usepackage[normalem]{ulem} 
\renewcommand\vec{\boldsymbol}

\usepackage{float}
\restylefloat{table}
\usepackage{tikz}
\usetikzlibrary{decorations.pathmorphing}
\usetikzlibrary{decorations.pathreplacing}
\usepackage{listings}
\crefname{lstlisting}{listing}{listings}
\Crefname{lstlisting}{Listing}{Listings}
\usepackage{tikz,pgfplots,pgfplotstable}
\usepgfplotslibrary{groupplots,statistics,fillbetween}
\pgfplotsset{compat=1.13}
\pgfplotsset{
    dne/.style 2 args={
        x filter/.append code={
            \edef\tempa{\thisrow{#1}}
            \edef\tempb{#2}
            \ifx\tempa\tempb
            \else
                
            \fi
        }
    }
}
\pgfplotsset{
    line and fill/.style={
        legend image code/.code={%
            \fill [fill=##1, opacity=0.2, draw=none] (0mm,-1ex) -- (0mm,1ex) -- (6mm,1ex) -- (6mm,-1ex) -- cycle;
            \draw [draw=##1,thick] (0mm,0mm) -- (6mm,0mm);
        }
    }
}
\newcommand{\linewitherrorfilter}[7]{
    \addplot [name path=pluserror,draw=none,no markers,forget plot] table [x={#2},y expr=\thisrow{#3}+\thisrow{#4},dne={#5}{#6}] {#1};
    \addplot [name path=minuserror,draw=none,no markers,forget plot] table [x={#2},y expr=\thisrow{#3}-\thisrow{#4},dne={#5}{#6}] {#1};
    \addplot [forget plot,fill=#7,opacity=0.2]
    fill between[on layer={},of=pluserror and minuserror];
    \addplot [#7,thick,no markers,dne={#5}{#6},legend image code/.code={\fill [fill=#7, opacity=0.2, draw=none] (0mm,-1ex) -- (0mm,1ex) -- (6mm,1ex) -- (6mm,-1ex) -- cycle; \draw [#7,thick] (0mm,0mm) -- (6mm,0mm);}] table [x={#2},y={#3}] {#1};
}

\newcommand*\swname{2D-VSR-Sim}

\begin{document}

\begin{frontmatter}
\title{Design, Validation, and Case Studies of \swname{}, an Optimization-friendly Simulator of 2-D Voxel-based Soft Robots}
\author{Eric Medvet}
\author{Alberto Bartoli}
\author{Andrea De Lorenzo}
\author{Stefano Seriani}
\address{Department of Engineering and Architecture, University of Trieste, Italy}

\begin{abstract} 
    Voxel-based soft robots (VSRs) are aggregations of soft blocks whose design is amenable to optimization.
    We here present a software, \swname{}, for facilitating research concerning the optimization of VSRs body and brain.
    The software, written in Java, provides consistent interfaces for all the VSRs aspects suitable for optimization and considers by design the presence of sensing, i.e., the possibility of exploiting the feedback from the environment for controlling the VSR.
    We experimentally characterize, from a mechanical point of view, the VSRs that can be simulated with \swname{} and we discuss the computational burden of the simulation.
    Finally, we show how \swname{} can be used to repeat the experiments of significant previous studies and, in perspective, to provide experimental answers to a variety of research questions.
\end{abstract}


\end{frontmatter}

\section{Motivation and significance}
\label{sec:motivation}
Soft robotics~\cite{rus2015design,kim2013soft} is a field of robotics that studies robots composed of soft materials.
Soft robots may be able to perform tasks which are hard for rigid robots~\cite{cheney2015evolving}, thanks to their soft nature which gives them ``infinite degrees of freedom''~\cite{chen2017soft}.
This potential comes at the cost of a larger effort required for designing the robot, both in its controller (brain) and its shape (body).
For this reason, optimization, possibly by means of meta-heuristics, is an effective way to address the design of soft robots.

A kind of soft robots that is particularly relevant from the perspective of optimization is the one of \emph{voxel-based soft robots} (VSRs)~\cite{hiller2012automatic}.
VSRs are composed of many simple soft blocks (called \emph{voxels}) that can change their volumes: the way voxels are assembled defines the \emph{body} of the VSR, whereas the law according to which voxels change their volume over the time defines the \emph{brain} of the VSR.
VSRs are even more suited to optimization than soft robots in general: on one hand, it is difficult to take into account the complex interactions that may emerge from the concurrent ``behavior'' of many small blocks; on the other hand, the very modular nature of VSRs permits to tackle the robot design from different points of view and at different scales.
Indeed, many applications of optimization to VSRs have been proposed, e.g., \cite{cheney2013unshackling,cheney2014evolved,cheney2015evolving,corucci2018evolving,kriegman2018morphological,talamini2019evolutionary,kriegman2019automated}.
Automation offered by optimization fostered research by allowing researchers and practitioners to focus more on the broad goal of optimization, rather than on the finer details about how to perform it.

In this paper, we present a software for experimenting with the optimization of VSRs that we called \swname{}.
We designed \swname{} focusing mainly on two key steps of optimization: what to optimize and towards which goal.
As a result, \swname{} offers a consistent interface
\begin{inparaenum}[(a)]
    \item to the different components (e.g., body, brain, specific mechanisms for control signal propagation) of a VSR which are suitable for optimization and 
    \item to the task the VSR is requested to perform (e.g., locomotion, grasping of moving objects).
\end{inparaenum}
For not posing bounds on which kinds of optimization techniques can be applied, we built \swname{} without making any assumption in these terms.
That is, \swname{} leaves researchers great freedom on how to optimize: different techniques, e.g., evolutionary computation or reinforcement learning, can be used on VSRs by researchers of different disciplines, e.g., robotics, artificial life, learning representations.

Some software frameworks originated from needs similar to ours, namely \cite{hiller2014dynamic} (later wrapped in Evosoro~\cite{kriegman2017simulating}) and~\cite{austin2019titan}.
Other frameworks could be used for modeling and simulating VSRs, e.g.,~\cite{hu2019difftaichi}, but operate at a much lower abstraction level and require a larger design effort to the researcher.
\swname{} differs from those frameworks because it offers an higher level of abstraction to the description of the VSRs that favors the task of defining what to optimize.
In particular, we included by design the possibility for the VSR to \emph{sense} the environment: that is, the controller under optimization can use as inputs the current velocities, accelerations, rotations, etc., of each of the voxels.
\swname{} allows the user (i.e., a researcher) to exploit those sensing abilities out-of-the-box, thereby saving the effort for modeling and implementing them in the simulation.
A recent study showed that sensing the environment may be beneficial for obtaining a broader range of behaviors~\cite{talamini2019evolutionary}.
Moreover, sensing might lead to a sharper arising of the \emph{embodied cognition paradigm}, according to which the complexity of the behavior of a robotic agent depends on both its brain and its body~\citep{pfeifer2006body}. 

Besides sensing, \swname{} differs from existing software tools also because it simulates a 2-D version of VSRs: operating in 2-D, rather than in 3-D, makes the search space in general ``smaller'' and hence potentially facilitates the optimization.
On the other hand, optimized artifacts have no clear real counterparts.
Indeed, some attempts to physically build VSRs are being made~\cite{hiller2012automatic,kriegman2019scalable,sui2020automatic,kriegman2020scalable}, at different scales and with different actuation mechanisms, but practicality is still limited.
We plan to extend \swname{} to the 3-D case as future work.

Finally, \swname{} provides components for \emph{visualizing} the simulated behavior of a VSR, which is very important in an exploratory research field of this kind.
Moreover, this functionality has been obtained by the separation of concerns design principle, by exploiting the programming language features offered by Java, which greatly simplifies possible extensions of \swname{}.

\section{Software description}
\label{sec:description}
The software here proposed is a simulator of one or more 2-D VSRs that perform a task, i.e., some activity whose degree of accomplishment can be evaluated quantitatively according to one or more indexes.
The simulation is discrete in time, using a fixed time-step, and continuous in space: the position and configuration of each voxel of the VSR is updated at each time-step according to the mechanical model and to the VSR controller.

\subsection{Voxel model}
\label{sec:voxel-model}
\begin{figure}[!h]
    \centering
    \begin{tikzpicture}
        \fill[gray] (0,0) rectangle (0.9,0.9);
        \fill[gray] (0,3) rectangle (0.9,2.1);
        \fill[gray] (3,0) rectangle (2.1,0.9);
        \fill[gray] (3,3) rectangle (2.1,2.1);
        \draw[blue,decorate,decoration={coil,amplitude=2pt,segment length=3pt}] (0,0.9) -- (0,2.1);
        \draw[blue,decorate,decoration={coil,amplitude=2pt,segment length=3pt}] (3,0.9) -- (3,2.1);
        \draw[blue,decorate,decoration={coil,amplitude=2pt,segment length=3pt}] (0.9,0) -- (2.1,0);
        \draw[blue,decorate,decoration={coil,amplitude=2pt,segment length=3pt}] (0.9,3) -- (2.1,3);
        \draw[red,decorate,decoration={coil,amplitude=2pt,segment length=3pt}] (0.9,0.9) -- (0.9,2.1);
        \draw[red,decorate,decoration={coil,amplitude=2pt,segment length=3pt}] (2.1,0.9) -- (2.1,2.1);
        \draw[red,decorate,decoration={coil,amplitude=2pt,segment length=3pt}] (0.9,0.9) -- (2.1,0.9);
        \draw[red,decorate,decoration={coil,amplitude=2pt,segment length=3pt}] (0.9,2.1) -- (2.1,2.1);
        \draw[green,decorate,decoration={coil,amplitude=2pt,segment length=3pt}] (0,0.9) -- (0.9,2.1);
        \draw[green,decorate,decoration={coil,amplitude=2pt,segment length=3pt}] (0.9,0.9) -- (0,2.1);
        \draw[green,decorate,decoration={coil,amplitude=2pt,segment length=3pt}] (3,0.9) -- (2.1,2.1);
        \draw[green,decorate,decoration={coil,amplitude=2pt,segment length=3pt}] (2.1,0.9) -- (3,2.1);
        \draw[green,decorate,decoration={coil,amplitude=2pt,segment length=3pt}] (0.9,0) -- (2.1,0.9);
        \draw[green,decorate,decoration={coil,amplitude=2pt,segment length=3pt}] (0.9,0.9) -- (2.1,0);
        \draw[green,decorate,decoration={coil,amplitude=2pt,segment length=3pt}] (0.9,3) -- (2.1,2.1);
        \draw[green,decorate,decoration={coil,amplitude=2pt,segment length=3pt}] (0.9,2.1) -- (2.1,3);
        \draw[orange,decorate,decoration={coil,amplitude=2pt,segment length=3pt}] (0.45,0.45) -- (2.55,2.55);
        \draw[orange,decorate,decoration={coil,amplitude=2pt,segment length=3pt}] (0.45,2.55) -- (2.55,0.45);
        \draw[black] (0.45,0.45) -- (2.55,2.55);
        \draw[black] (0.45,2.55) -- (2.55,0.45);
        \draw [decorate,decoration={brace,amplitude=10pt},xshift=-4pt,yshift=0pt] (-0.25,0) -- (-0.25,3) node [black,midway,xshift=-0.6cm] {$l$};
        \draw [decorate,decoration={brace,amplitude=10pt,mirror},xshift=0pt,yshift=-4pt] (0,-0.25) -- (3,-0.25) node [black,midway,yshift=-0.6cm] {$l$};
        \draw [decorate,decoration={brace,amplitude=5pt},xshift=0pt,yshift=2pt] (2.1,3.25) -- (3,3.25) node [black,midway,yshift=0.6cm] {$l_m l$};
        \draw [decorate,decoration={brace,amplitude=5pt},xshift=2pt,yshift=0pt] (3.25,3) -- (3.25,2.1) node [black,midway,xshift=0.6cm] {$l_m l$};
        \node at (0.2, 0.2) {$4$};
        \node at (2.8, 0.2) {$3$};
        \node at (0.2, 2.8) {$1$};
        \node at (2.8, 2.8) {$2$};
    \end{tikzpicture}
    \caption{
        The mechanical model of the voxel.
        The four masses are depicted in gray and numbered (for later reference) in black; the different components of the scaffolding are depicted in blue, green, red, and orange (see text); the ropes are depicted in black.
    }
    \label{fig:voxel-model}
\end{figure}
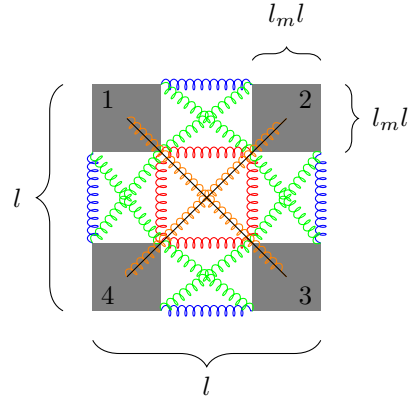

A voxel is a soft 2-D block, i.e., a deformable square modeled with four rigid bodies (square \emph{masses}), a number of spring-damper systems (SDSs) that constitute a \emph{scaffolding}, and a number of \emph{ropes}.
SDSs and ropes have zero mass: ropes act as upper bounds to the distance that two bodies can have.
\Cref{fig:voxel-model} shows the mechanical model of a single voxel.
Most of the properties of the voxel model are \emph{configurable} by the user, as explained below.

The user can configure the scaffolding specifying a subset of the following groups of SDSs:
\begin{inparaenum}[(a)]
    \item \emph{side external}, one outer SDS connecting the two masses for each voxel side (blue in \Cref{fig:voxel-model});
    \item \emph{side internal}, one inner SDS connecting the two masses for each voxel side (red in \Cref{fig:voxel-model});
    \item \emph{side cross}, two crossing SDSs connecting the two masses for each voxel side (green in \Cref{fig:voxel-model});
    \item \emph{central cross}, two crossing SDSs connecting the four masses (orange in \Cref{fig:voxel-model}).
\end{inparaenum}
The presence of the ropes can be configured (enabled or disabled) by the user too.
\Cref{tab:voxel-props} shows the main parameters of the voxel model along with their default values and domains.
Mass friction and restitution coefficients are used by the physics engine (see \Cref{sec:architecture-functionalities}) while determining the effects of collisions of masses with other bodies (e.g., the ground).

\begin{table}[!h]
    \centering
    \begin{tabular}{l S[table-format=2.1] c l}
        \toprule
        Description and symbol & {Def.\ val.} & Domain & Unit\\
        \midrule
        Side length $l$ & 3 & $\left]0,+\infty\right[$ & \si{\meter} \\
        Mass side length ratio $l_m$ & 0.3 & $\left]0,0.5\right]$ &  \\
        Mass linear damping $d^l_m$ & 1 & $\left[0,+\infty\right[$ &  \\
        Mass angular damping $d^\omega_m$ & 1 & $\left[0,+\infty\right[$ &  \\
        Mass mass $m_m$ & 1 & $\left]0,+\infty\right[$ & \si{\kilo\gram} \\
        Mass friction coeff.\ $\rho_m$ & 100 & $\left[0,+\infty\right[$ & \\
        Mass restitution coeff.\ $r_m$ & 0.1 & $\left]0,+\infty\right[$ & \\
        SDS frequency $f_s$ & 8 & $\left[0,+\infty\right[$ & \si{\hertz} \\
        SDS damping ratio $d_s$ & 0.3 & $[0,1]$ & \\
        Max force magn.\ $f_{\max}$ & 1000 & $\left[0,+\infty\right[$ & \si{\newton} \\
        Max area change $\rho_A$ & 0.2 $[0,1]$ & \\
        \bottomrule
    \end{tabular}
    \caption{Voxel configurable properties.}
    \label{tab:voxel-props}
\end{table}

By varying the values of the parameters, the user can impact on the properties of the material constituting the voxel.
In particular, for impacting on the softness of the voxel, the user can operate on the scaffolding and/or on the SDS frequency $f_s$; with the former, the more the selected groups, the more rigid the voxel.
After a few exploratory experiments, we set as default value $f_s=\SI{8}{\hertz}$ and the scaffolding composed of all the groups.

The VSR can perform its task by varying the area of the composing voxels over the time, i.e., by actuating each voxel.
In the mechanical model of \swname{}, the actuation can be obtained either by imposing forces on the four masses, or by varying the resting length of the SDSs---the method to be used can be configured by the user.
In the first case (\emph{force actuation mode}), the force vector for each mass is applied on the mass center and along the direction connecting the mass center with the voxel center (i.e., the average of the centers of the four masses).
The magnitude of the force vector is $f f_{\max}$, where $f_{\max}$ is a maximum configurable value and $f \in [-1,1]$ is the control value, $f=1$ corresponding to shrinking the voxel.
In the second case (\emph{area actuation mode}), the resting length of all the SDSs is instantaneously modified such that the voxel side becomes $l' = \sqrt{l^2 (1-f \rho_A)}$ where $\rho_A$ is the maximum increase or decrease of the voxel area that is meant to be obtained by applying a control value $f \in [-1,1]$.
Note, however, that the actual area depends also on the other forces acting on the voxel, namely those related to gravity, adjacent connected voxels, or static objects (like the ground).
We set as default actuation method the second one.

\subsection{VSR model}
\label{sec:vsr-model}
\begin{figure*}
    \centering
    \def\voxel at (#1,#2) {
        \fill[gray] (#1+0,#2+0) rectangle (#1+0.9,#2+0.9);
        \fill[gray] (#1+0,#2+3) rectangle (#1+0.9,#2+2.1);
        \fill[gray] (#1+3,#2+0) rectangle (#1+2.1,#2+0.9);
        \fill[gray] (#1+3,#2+3) rectangle (#1+2.1,#2+2.1);
        \draw[gray,decorate,decoration={coil,amplitude=2pt,segment length=3pt}] (#1+0,#2+0.9) -- (#1+0,#2+2.1);
        \draw[gray,decorate,decoration={coil,amplitude=2pt,segment length=3pt}] (#1+3,#2+0.9) -- (#1+3,#2+2.1);
        \draw[gray,decorate,decoration={coil,amplitude=2pt,segment length=3pt}] (#1+0.9,#2+0) -- (#1+2.1,#2+0);
        \draw[gray,decorate,decoration={coil,amplitude=2pt,segment length=3pt}] (#1+0.9,#2+3) -- (#1+2.1,#2+3);
        \draw[gray,decorate,decoration={coil,amplitude=2pt,segment length=3pt}] (#1+0.45,#2+0.45) -- (#1+2.55,#2+2.55);
        \draw[gray,decorate,decoration={coil,amplitude=2pt,segment length=3pt}] (#1+0.45,#2+2.55) -- (#1+2.55,#2+0.45);
        \draw[gray] (#1+0.45,#2+0.45) -- (#1+2.55,#2+2.55);
        \draw[gray] (#1+0.45,#2+2.55) -- (#1+2.55,#2+0.45);
    }
    \begin{tikzpicture}
        \voxel at (0,3)
        \voxel at (3,3)
        \voxel at (6,3)
        \voxel at (9,3)
        \voxel at (0,0)
        \voxel at (9,0)
        \draw[red,thick] (2.55,3.45) -- (3.45,3.45);
        \draw[red,thick] (2.55,5.55) -- (3.45,5.55);
        \draw[red,thick] (5.55,3.45) -- (6.45,3.45);
        \draw[red,thick] (5.55,5.55) -- (6.45,5.55);
        \draw[red,thick] (8.55,3.45) -- (9.45,3.45);
        \draw[red,thick] (8.55,5.55) -- (9.45,5.55);
        \draw[red,thick] (0.45,2.55) -- (0.45,3.45);
        \draw[red,thick] (2.55,2.55) -- (2.55,3.45);
        \draw[red,thick] (9.45,2.55) -- (9.45,3.45);
        \draw[red,thick] (11.55,2.55) -- (11.55,3.45);
    \end{tikzpicture}
    \caption{
        The mechanical model of a VSR composed of $8$ voxels.
        The masses, the SDSs, and the ropes are depicted in gray; the rigid joints connecting the voxels are depicted in red.
    }
    \label{fig:vsr-model}
\end{figure*}
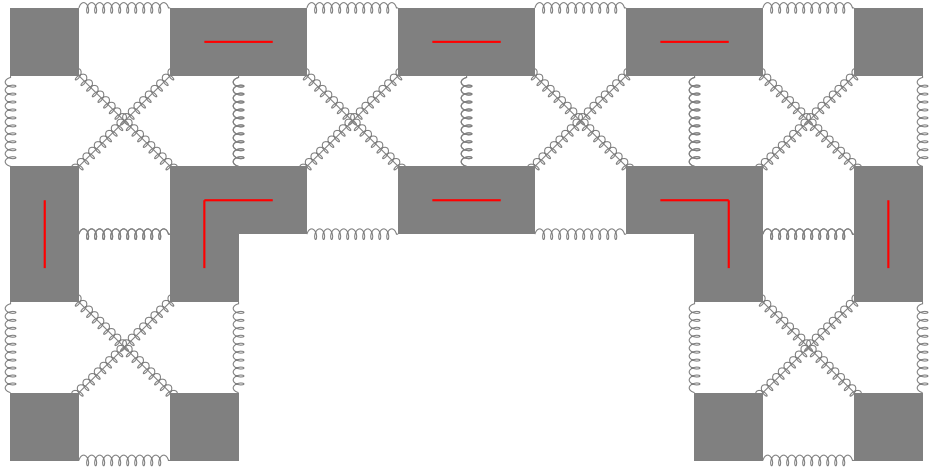

A VSR is modeled as a collection of voxels organized in a 2-D grid, each voxel in the grid being rigidly connected with the voxel above, below, on the left, and on the right.
The connection between two voxels is modeled as two rigid joints connecting the centers of the masses on the common side.
The rigid joint does not allow rotations of the masses around the connection points, nor variation in the distance between the two connected masses: in other words, two masses connected by a rigid joint are welded.
\Cref{fig:vsr-model} shows the mechanical model of an example VSR composed of $8$ voxels.

The parameters of the voxels composing a VSR can have different values (with the exception of the side length $l$); a VSR can hence be composed of different materials.

\subsection{VSR controller}
\label{sec:vsr-controller}
The way a VSR behaves is determined by a \emph{controller}.
Whenever it is invoked, the controller determines, for each voxel $v_i$ of the VSR, the control value $f_i \in [-1,1]$ to apply.
The control value is applied by the physics engine (see \Cref{sec:simulation}) and results in a change in the area of the corresponding voxel and hence a change in the shape of the VSR.
The controller can be implemented by the user and \swname{} provides ample freedom in this respect: realizable controllers include those where $f_i$ depends only on the current time $t$ and those where $f_i$ is the result of a possibly non trivial processing of current and previous states of the VSR and the environment.

A controller has access to several \emph{sensors} for each voxel, hence giving the VSR the ability to sense its status and the environment.
For each sensor $s$ and each voxel $v_i$, the controller may use zero or more of the following:
\begin{inparaenum}[(a)]
    \item the current value $s(t, v_i)$;
    \item the average value $\frac{1}{n} \sum_{k=0}^{n-1} s(t-k \Delta t_c, v_i)$ of the last $n$ readings (at times $t, \dots, t-(n-1) \Delta t_c$);
    \item the $n$-th difference $s(t, v_i)-s(t-(n-1) \Delta t_c, v_i)$ between the current value and the $n-1$-th reading.
\end{inparaenum}
Available sensors include:
\begin{inparaenum}[(a)]
    \item \emph{area ratio}, i.e., the ratio between the current area of the voxel and its rest area $l^2$;
    \item \emph{velocity magnitude}, the magnitude $\|\vec{v}\|$ of the voxel velocity $\vec{v}$ obtained as the average velocity of the four masses;
    \item \emph{angle}, the rotation angle $\alpha$ of the voxel obtained as the average of the direction of the vector connecting masses $1$ and $2$ and that of the vector connecting masses $4$ and $3$ (see \Cref{fig:voxel-model});
    \item \emph{$x$-velocity}, the velocity $v_x = \vec{v} \cdot (1,0)$ along the $x$-axis;
    \item \emph{$y$-velocity}, the velocity $v_y = \vec{v} \cdot (0,1)$ along the $y$-axis;
    \item \emph{rotated $x$-velocity}, the velocity $v^\text{rot}_x = \vec{v} \cdot (\cos \alpha, \sin \alpha)$ along the $x$-axis rotated by $\alpha$;
    \item \emph{rotated $y$-velocity}, the velocity $v^\text{rot}_y = \vec{v} \cdot (\sin \alpha, -\cos \alpha)$ along the $y$-axis rotated by $\alpha$;
    \item \emph{touching}, a value in $\{0,1\}$ that is $1$ if and only if at least one of the masses is currently in touch with another body not belonging to the VSR.
\end{inparaenum}

We implemented two controllers in \swname{}, one being stateless and not exploiting any sensor, the other based on a multilayer perceptron (MLP) and resembling the one presented in~\cite{talamini2019evolutionary}.
The two implementations can be used without writing any code and can be instantiated by setting the values of the parameters of the corresponding controllers, that we describe below.

The stateless, non-sensing controller simply applies to each voxel $v_i$ a control value $f_i=f_i(t)$: the controller parameters are hence a grid of functions $f_i: \mathbb{R^+} \to [-1,1]$.

The MLP-based controller computes, at each invocation, the output $\vec{y} = f_\text{MLP}(\vec{x}, \vec{\theta})$ of a MLP with a user-defined architecture and weights $\vec{\theta}$ on an input $\vec{x}$.
The size of the input layer $m, \vec{x} \in \mathbb{R}^m$ is implicitly defined by the user-defined sensors for each voxel of the VSR; the size of the output layer $n, \vec{y} \in [-1,1]^n$ is equal to the number of voxels of the VSR.
The control value $f_i$ of a voxel $v_i$ is set to the $i$-th element $y_i$ of $\vec{y}$.
Optionally, a supplementary input can be set to the current value of a user-defined function of the current time, called driving function.
The MLP-based controller parameters are hence: the grid of sensors to be used, the driving function, the MLP architecture, and the MLP weights $\vec{\theta}$.
For example, let the VSR be composed of $6$ voxels (like the one of \Cref{fig:vsr-model}) and let the inputs be: the average of the last $5$ area ratio, rotated $x$-velocity, and rotated $y$-velocity readings for all the voxels; the current touching reading for the $2$ bottom voxels; a driving function $\sin(2 \pi t)$.
Let the MLP consist of the inner layer, an hidden layer of $10$ neurons, and the output layer.
Then, the number $|\vec{\theta}|$ of weights is given by $(6 \times 3+2+1+1) \times 10 + 10 \times 6 = 220 + 60 = 280$: $6$ voxels with $3$ sensors, $2$ voxels with one sensor each, $1$ driving function, the bias.
That vector might be the one subjected to optimization (see \Cref{sec:example-mlp}).

\subsection{Simulation}
\label{sec:simulation}
\swname{} exploits an existing physics engine, dyn4j\footnote{\url{http://www.dyn4j.org/}}, for solving the mechanical model defined by a VSR subjected to the forces caused by the actuation determined by its controller and by the interaction with other bodies (typically, the ground).
We refer the reader to the documentation of dyn4j for the fine details about how physics is modeled.

The physics engine can be configured in several aspects.
One that is particularly relevant is the time-step $\Delta t$ that is used for numerically solving the model: in order to avoid numerical instabilities, $\Delta t$ has to be small enough with respect to the SDS frequency $f_s$.
For the same reason, we made \swname{} to not invoke the controller of the VSR at every time-step, but every $c+1$ time-steps, $c \ge 0$ being the \emph{control step interval}.
After preliminary experiments, we set the default values for $\Delta t$ to \SI[quotient-mode=fraction]{1/60}{\second} (which is the default value of the underlying physics engine) and for $c$ to $1$, i.e., the controller is invoked every \SI[quotient-mode=fraction]{1/30}{\second}.

\subsection{Task}
The goal of the optimization is represented in \swname{} as a \emph{task} to be solved.
In general, a task is a function that processes an input and gives an output: the common case is the one where the input is a description of a VSR (or of part of it) and the output is a measure of the degree to which that VSR accomplished the task.
The latter can be based on any quantity that made available by \swname{} and the underlying physics engine, e.g., position of the center of mass of the VSR, consumed energy, and so on.

We implemented one task in \swname{} that represents \emph{locomotion}, i.e., a task where the goal of the robot is to travel as far as possible.
We chose this task because it is the one on which the vast majority of previous studies put their focus.
The implementation of this task can be used without writing any code and can be instantiated by setting the values of the parameters describe below.

The input of this task is a description of the VSR, in terms of its body (grid of voxels described by their parameters) and brain (controller).
The output is given by one or more measures obtained by simulating the VSR that moves on the ground as, e.g., the distance traveled along the $x$-direction.
Many aspects of the locomotion task can be configured, including:
\begin{inparaenum}[(a)]
    \item the duration of the simulation;
    \item the shape of the ground, that can be flat or uneven with different degrees of roughness;
    \item the measures to be taken upon the completion of the simulation.
\end{inparaenum}
The latter aspect allows to cast the optimization of a VSR for locomotion as a single- or multi-objective optimization problem.
Measures that may be taken include:
\begin{inparaenum}[(a)]
    \item \label{item:locomotion-obj-distance} the travel velocity, i.e., the ratio between the displacement of the VSR center (the average of its voxels centers) along the $x$-axis and the duration of the simulation;
    \item \label{item:locomotion-obj-ycenter} the average $y$-coordinate of the VSR center;
    \item \label{item:locomotion-obj-control} the average of the sum of the squared control values of the voxels.
\end{inparaenum}
For example, optimizing the controller of a VSR with a given body for the minimization of both measure \ref{item:locomotion-obj-distance} (inverted in sign) and measure \ref{item:locomotion-obj-control} corresponds to searching for a controller that is at the same time good in running and energy-saving.

\subsection{Software architecture}
\label{sec:architecture-functionalities}
\swname{} is meant to be used within or together with another software that performs the actual optimization.

\swname{} is organized as a Java package containing the classes and the interfaces that represent the models and concepts described previously.
The voxel is represented by the \lstinline{Voxel} class and its parameters can be specified using the \lstinline{Voxel.Builder} class, using the builder pattern.
The VSR is represented by the \lstinline{VoxelCompound} class; a description of a VSR, that can be used for building a VSR accordingly, is represented by the \lstinline{VoxelCompound.Description} class.
A controller is represented by the interface \lstinline{Controller}: a controller exploiting sensing can be realized by extending the abstract class \lstinline{ClosedLoopController}, that takes care of collecting the sensor readings and makes them available to the inheriting class.
Finally, a task is described by the interface \lstinline{Task}.

Since VSRs are defined as grids of voxels, a class of particular importance in \swname{} is \lstinline{Grid<T>}, that represents a 2-D grid of objects of type \lstinline{T}: \lstinline{Grid} may contain null objects, meaning that the corresponding positions are empty.
For example, a \lstinline{Grid<Voxel.Builder>} is used for specifying the body of a VSR, whereas a \lstinline{Grid<SerializableFunction<Double, Double>>} is used for specifying the time functions, one for each voxel, that are the parameters of the stateless, non-sensing controller (represented by the class \lstinline{TimeFunction}).

\swname{} provides a mechanism for keeping track of an ongoing simulation based on the observer pattern.
A \lstinline{SnapshotListener} interface represents the observer that is notified of progresses in the simulation, each in the form of a \lstinline{Snapshot}: the latter is an immutable representation of the state of all the objects (e.g., positions of voxels, values of their sensor readings) in the simulation at a given time.
We implemented two listeners using this interface.
\lstinline{GridOnlineViewer} renders a visualization of the simulated world within a GUI, whereas \lstinline{GridFileWriter} produces a video file: both can process multiple simulations together, organized in a grid.
The possibility of visualizing many simulations together can be useful, for example, for comparing different stages of an optimization.

\begin{figure*}[!h]
    \centering
    \includegraphics[width=1\linewidth]{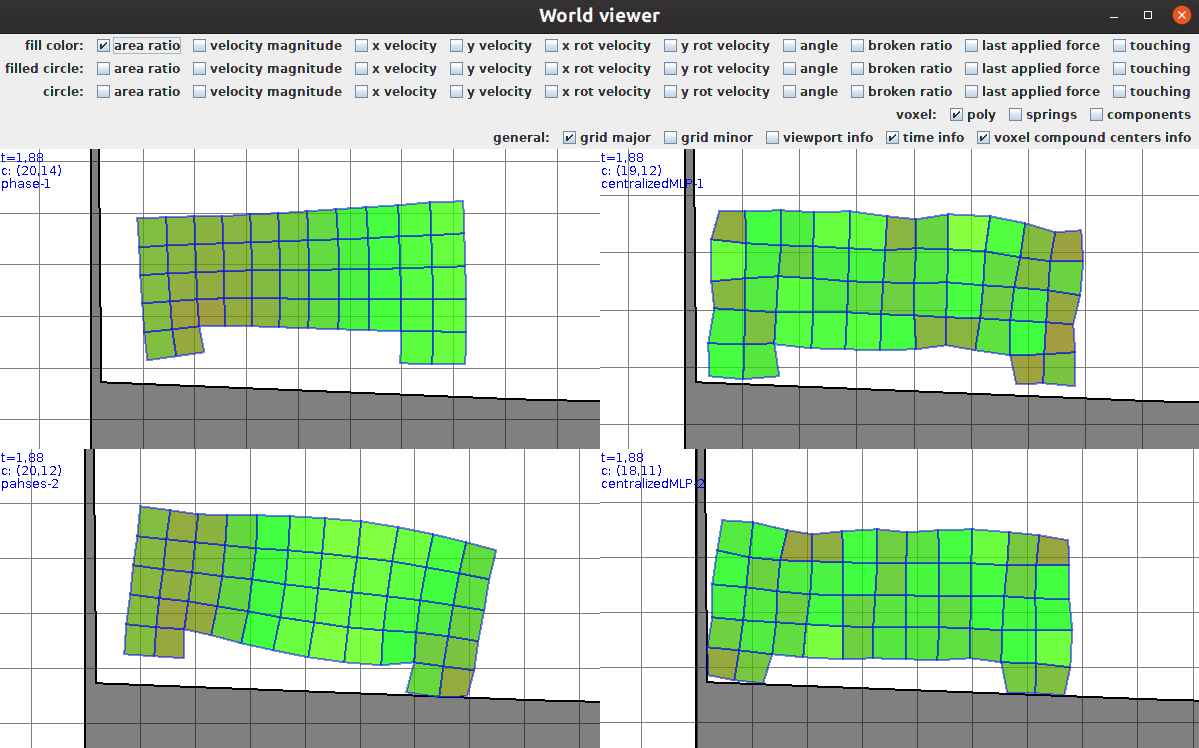}
    \caption{
        The GUI provided by \swname{} for visualizing many simulations together (here four simulations in a grid of $2 \times 2$).
        In this (default) configuration, each voxel is filled with a color that is proportional to the variation of the voxel area: red means shrunk, green means no variation, yellow means enlarged.
        The ground is filled in dark gray.
    }
    \label{fig:gui}
\end{figure*}

\Cref{fig:gui} shows the graphical user interface (GUI) provided by \lstinline{GridOnlineViewer}: four simulations of the locomotion task are shown with four different VSRs.
On top of the GUI, a set of checkboxes allows the user to customize the visualization with immediate effect.
Several measures can be visualized through the fill color of the voxel or other means; voxels SDSs and masses can be visualized too, as well as other useful information.

\subsection{Sample code}
\label{sec:sample-code}
In \Cref{lst:example}, we show a brief fragment of Java code using \swname{} for setting up a VSR and assessing it in the task of locomotion.

\begin{lstlisting}[
    caption={
        Example of usage of \swname{} for setting up and performing a simulation of a VSR composed of two materials that performs locomotion.
    },
    basicstyle=\ttfamily\footnotesize,
    breaklines=true,
    label={lst:example},
    escapechar=|
]
Locomotion locomotion = new Locomotion( |\label{code-line:task-def-start}|
  60,
  Locomotion.createTerrain("flat"),
  Lists.newArrayList(
    Locomotion.Metric.TRAVEL_X_VELOCITY,
    Locomotion.Metric
      .AVG_SUM_OF_SQUARED_CONTROL_SIGNALS
  ),
  1,
  new Settings()
); |\label{code-line:task-def-end}|
final Voxel.Builder hardMaterial = Voxel.Builder.create() |\label{code-line:material-1-def-start}|
  .springF(25)
  .springScaffoldings(EnumSet.allOf(
    Voxel.SpringScaffolding.class)); |\label{code-line:material-1-def-end}|
final Voxel.Builder softMaterial = Voxel.Builder.create() |\label{code-line:material-2-def-start}|
  .springF(5)
  .springScaffoldings(EnumSet.of(
    Voxel.SpringScaffolding.SIDE_EXTERNAL,
    Voxel.SpringScaffolding.CENTRAL_CROSS)); |\label{code-line:material-2-def-end}|
int w = 4;
int h = 2;
VoxelCompound.Description description = new VoxelCompound.Description( |\label{code-line:vsr-def-start}|
  Grid.create( |\label{code-line:vsr-body-def-start}|
    w, h,
    (x, y) -> (y == 0) ? materialHard : materialSoft
  ), |\label{code-line:vsr-body-def-end}|
  new TimeFunction(Grid.create( |\label{code-line:vsr-brain-def-start}|
    w, h,
    (x, y) -> (Double t) -> Math.sin(-2 * Math.PI * t + Math.PI * ((double) x / (double) w))
  )) |\label{code-line:vsr-brain-def-end}|
); |\label{code-line:vsr-def-end}|
List<Double> result = locomotion.apply(description); |\label{code-line:task-application}|
\end{lstlisting}

From line~\ref{code-line:task-def-start} to line~\ref{code-line:task-def-end}, a task of locomotion on flat terrain is defined with duration a of \SI{60}{\second}, a control step interval $c=1$, default settings for the physics engine, and two measures to be taken at the end of the simulation: the velocity along the $x$-axis, and average sum of the squared control values.

From line~\ref{code-line:material-1-def-start} to line~\ref{code-line:material-2-def-end} two materials are defined by instantiating two \lstinline{Voxel.Builder} objects with different parameters (using the builder pattern).
The first material (\lstinline{hardMaterial}, lines \ref{code-line:material-1-def-start}--\ref{code-line:material-1-def-end}) has a large SDS frequency $f_s=\SI{25}{\hertz}$ and all the groups of SDSs are used in the scaffolding.
The second material (\lstinline{softMaterial}, lines \ref{code-line:material-2-def-start}--\ref{code-line:material-2-def-end}) has $f_s=\SI{5}{\hertz}$ and only side external and central cross groups enabled.

From line~\ref{code-line:vsr-def-start} to line~\ref{code-line:vsr-def-end} a VSR is defined by defining the corresponding \lstinline{VoxelCompound.Description}.
The definition consists of a desription of the body (lines \ref{code-line:vsr-body-def-start}--\ref{code-line:vsr-body-def-end}) and a description of the brain (lines \ref{code-line:vsr-brain-def-start}--\ref{code-line:vsr-brain-def-end}).
The former is defined as a grid of $4 \times 2$ \lstinline{Voxel.Builder}s: the bottom row of the grid uses the hard material; the top row uses the soft material.
The brain, i.e., the controller, is the stateless, non-sensing one (\lstinline{TimeFunction}) in which the control value depends only on the current time.
In particular, for a given voxel at position $x,y$ in the grid, the control value is given by $f_{x,y}(t)=\sin\left(-2 \pi t + \pi \frac{x}{4}\right)$---i.e., the control value varies with the same frequency and amplitude, but differs in phase along the body.

Finally, at line~\ref{code-line:task-application}, the execution of the task by the VSR is simulated.
This is the statement that, when executed, results in the actual computation of the metrics that are, in \Cref{lst:example}, eventually stored in \lstinline{result} in the form of a list of two numbers.

It can be seen that, by exploiting the recent Java language features (namely, lambda expressions) and established coding practices, \swname{} allows to specify complex scenarios concisely and consistently.

\subsection{Impact of main parameters}
\label{sec:mechanical-validation}
We experimentally characterized \swname{} in terms of the impact of its main parameters.
In particular, we considered three parameters---the SDS frequency $f_s$, the scaffolding, and the time-step $\Delta t$---and evaluated their impact on the static and dynamic behavior of the simulated material and on the simulation performance.

\subsubsection{Static and dynamic behavior}
\begin{figure}[!h]
    \centering
    \includegraphics[width=1\linewidth]{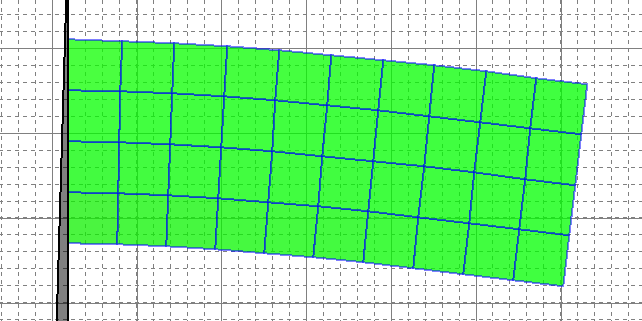}
    \caption{
        The $10 \times 4$ cantilever beam used for the characterization of the static and dynamic behavior of the simulated material.
    }
    \label{fig:cantilever}
\end{figure}

For characterizing the static and dynamic behavior, we considered a cantilever beam, i.e., a grid of $w \times h$ voxels with the same properties in which the $h$ leftmost voxels are connected with an unamovable body (a wall).
We experimented with three cantilever sizes ($5 \times 2$, $10 \times 2$, and $10 \times 2$)---\Cref{fig:cantilever} shows the $10 \times 4$ case.

We simulated an interval of \SI{60}{\second} with the default values for all the parameters with the exception of the SDS frequency $f_s$ and the scaffolding.
We set $f_s$ to one value in \SIlist{4;8;10;15;20;25;30}{\hertz} and scaffolding to one possible subset of the scaffolding groups---we varied only one parameter at once.
During the simulation, we applied a force $\vec{f}$ along the $y$-axis, i.e., towards the ground, to the rightmost extreme of the cantilever beam: in detail, we applied an equal force with a magnitude of $\frac{1}{2h} \|\vec{f}\|$ to each one of the two rightmost masses of each of the $h$ rightmost voxels.

For characterizing the static behavior, we set $\|\vec{f}\|=\SI{30}{\newton}$ and applied it for the entire duration of the simulation.
For the dynamic case, we set $\|\vec{f}\|=\SI{800}{\newton}$ and applied it for the initial \SI{0.1}{\second}.
In both cases, we disabled the gravity.
We measured the $y$-displacement of the center of the $h$ rightmost voxels with respect to the resting position.

\begin{figure*}
    \centering
    \begin{tikzpicture}
        \begin{groupplot}[
            width=.5\linewidth,
            height=.3\linewidth,
            grid=both,
            ymin=-6.3,ymax=-5.95,
            group style={
                group size=2 by 1,
                horizontal sep=3mm,
                xticklabels at=edge bottom,
                yticklabels at=edge left
            }
        ]
            \nextgroupplot[
                xlabel={$f_s$ [\si{\hertz}]},
                ylabel={Final $y$-displacement [\si{\meter}]},
                legend columns=1,
                legend entries={$5 \times 2$, $10 \times 2$, $10 \times 4$},
                legend pos=south east
            ]
            \addplot table [x=builder.springF,y=yDisplacement,dne={shape}{10x4}] {data/cantilever-static-fs.txt};
            \addplot table [x=builder.springF,y=yDisplacement,dne={shape}{15x4}] {data/cantilever-static-fs.txt};
            \addplot table [x=builder.springF,y=yDisplacement,dne={shape}{20x4}] {data/cantilever-static-fs.txt};
            \nextgroupplot[
                xlabel={Spring scaffolding groups},
                legend columns=1,
                legend entries={$5 \times 2$, $10 \times 2$, $10 \times 4$},
                legend pos=south east,
                symbolic x coords={E+C,E+I+C,All,E+I+X}
            ]
            \addplot table [only marks,x=builder.springScaffoldings,y=yDisplacement,dne={shape}{10x4}] {data/cantilever-static-scaff.txt};
            \addplot table [only marks,x=builder.springScaffoldings,y=yDisplacement,dne={shape}{15x4}] {data/cantilever-static-scaff.txt};
            \addplot table [only marks,x=builder.springScaffoldings,y=yDisplacement,dne={shape}{20x4}] {data/cantilever-static-scaff.txt};
        \end{groupplot}
    \end{tikzpicture}
    \caption{
        Mechanical characterization in static conditions: final $y$-displacement of a cantilever beam subjected to a constant force for three shapes.
        On the left, vs.\ the SDS frequency $f_s$ (with the default value for all the other parameters); on the right, vs.\ different configurations of the scaffolding---E, I, X, and C denoting, respectively side external, side internal, side cross, and central cross.
    }
    \label{fig:cantilever-static}
\end{figure*}
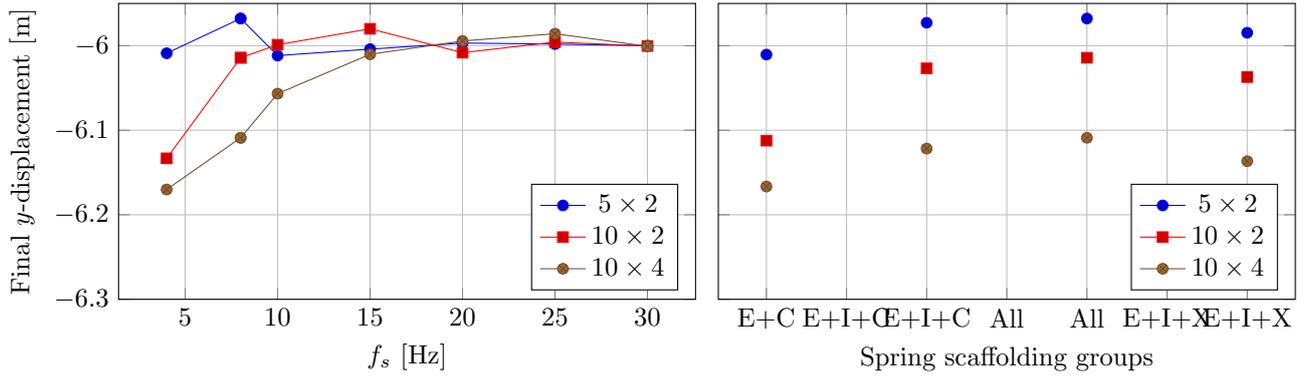

\Cref{fig:cantilever-static} shows the results for the static case in terms of the final $y$-displacement at $t=\SI{60}{\second}$ vs.\ the parameter value (SDS frequency $f_s$ on the left plot and scaffolding on the right plot), one set of measurements for each of the three cantilever sizes.
It can be seen that the static behavior is, in general, consistent.
The lower $f_s$, the larger the absolute $y$-displacement: i.e., the softer the material, the more the cantilever bends.
The same trend can be observed for the scaffolding: with just two groups (side external and central cross) the $y$-displacement is larger than with more groups; as expected, the lowest displacement is obtained with all the groups enabled, corresponding to the largest stiffness.

\begin{figure*}[!h]
    \centering
    \begin{tikzpicture}
        \begin{groupplot}[
            width=.5\linewidth,
            height=.3\linewidth,
            grid=both,
            ymin=-10,ymax=2,
            xlabel={Time [\si{\second}]},
            no markers,
            group style={
                group size=2 by 1,
                horizontal sep=3mm,
                xticklabels at=edge bottom,
                yticklabels at=edge left
            }
        ]
            \nextgroupplot[
                ylabel={$y$-displacement [\si{\meter}]},
                legend columns=1,
                legend entries={$\SI{3}{\hertz}$,$\SI{8}{\hertz}$,$\SI{15}{\hertz}$,$\SI{30}{\hertz}$},
                legend pos=south east
            ]
            \addplot table [x=st,y=y,dne={fs}{4}] {data/cantilever-dynamic-fs.txt};
            \addplot table [x=st,y=y,dne={fs}{8}] {data/cantilever-dynamic-fs.txt};
            \addplot table [x=st,y=y,dne={fs}{15}] {data/cantilever-dynamic-fs.txt};
            \addplot table [x=st,y=y,dne={fs}{30}] {data/cantilever-dynamic-fs.txt};
            \nextgroupplot[
                legend columns=1,
                legend entries={E+C,E+I+C,All,E+I+X},
                legend pos=south east
            ]
            \addplot table [x=st,y=y,dne={scaff}{ec}] {data/cantilever-dynamic-scaff.txt};
            \addplot table [x=st,y=y,dne={scaff}{eic}] {data/cantilever-dynamic-scaff.txt};
            \addplot table [x=st,y=y,dne={scaff}{eixc}] {data/cantilever-dynamic-scaff.txt};
            \addplot table [x=st,y=y,dne={scaff}{eix}] {data/cantilever-dynamic-scaff.txt};
        \end{groupplot}
    \end{tikzpicture}
    \caption{
        Mechanical characterization in dynamic conditions: $y$-displacement vs.\ time for a $10 \times 4$ cantilever beam subjected to an initial force impulse.
        On the left, for different values of the SDS frequency $f_s$ (with the default value for all the other parameters); on the right, for different configurations of the scaffolding.
    }
    \label{fig:cantilever-dynamic}
\end{figure*}
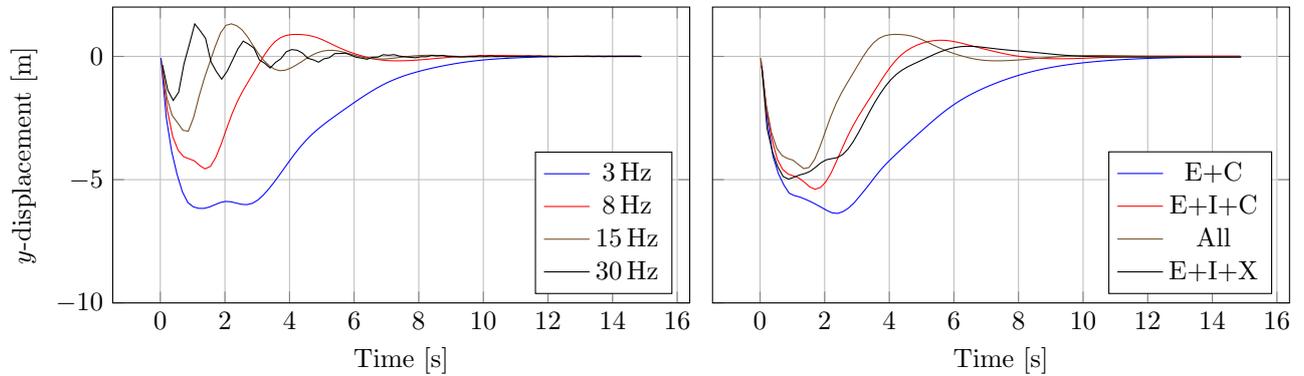

\Cref{fig:cantilever-dynamic} shows the results for the dynamic case in terms of the $y$-displacement over the time (for $t \le \SI{15}{\second}$) for different values of the SDS frequency $f_s$ (left plot) and different configurations of the scaffolding (right plot)---only the results for the $10 \times 4$ cantilever are shown.
It can be seen that the two parameters impact again as expected:
the lower $f_s$, hence the softer the material, the larger the maximum absolute $y$-displacement and the slower the oscillation.
Similarly, with all the groups the maximum displacement is smaller and the oscillation shorter than with just two groups.

\subsubsection{Simulation performance}
\label{sec:performance-validation}
Concerning the simulation performance, i.e., how many simulation steps can be performed in the unit of time on a given computing machine, we explored the impact of scaffolding and simulation step $\Delta t$.
The SDS frequency $f_s$ has no significant impact; we hence do not discuss this parameter here.

We considered a VSR of $w \times 3$ voxels with the same properties (i.e., a sort of worm of length $w$) actuated by a stateless, non-sensing controller.
The control value for a voxel at position $x,y$ in the grid is given by $f_{x,y}(t)=\sin\left(-2 \pi t + \pi \frac{x}{w} \right)$, i.e., the same of \Cref{sec:sample-code}.
The VSR moved on an uneven surface in the attempt of performing the locomotion task.

For each value of $w \in \{3, 6, \dots, 42, 45\}$ (starting from $w=45$), we performed $5$ simulations lasting \SI{60}{\second} (simulated time).
We executed the simulations in parallel (as \lstinline{Callable}s through the Java \lstinline{ExecutorService} framework, one \lstinline{Callable} for each core) on the Galileo partition of the CINECA HPC cluster, where each node is equipped with $2 \times 18$ cores based on \SI{2.30}{\giga\hertz} Intel Xeon E5-2697 v4 (Broadwell) and with \SI{128}{\giga\byte} RAM.
We used OpenJDK 64-Bit Server VM (build 13+33) with the \texttt{-Xmx8G} option (i.e., with at most \SI{8}{\giga\byte}) and repeated the procedure $10$ times on $10$ different HPC nodes, hence performing $5 \times 10=50$ simulations for each value of $w$.
At the end of each simulation, we counted the average number of \emph{simulated voxel steps per second (SVSPS)}, obtained as $\left(\frac{60}{\Delta t} 3 w \right) \frac{1}{\tau}$, $\frac{60}{\Delta t}$ being the number of simulated steps, $3 w$ the number of voxels, and $\tau$ the duration (wall time) in seconds of the simulation.

\begin{figure*}[!h]
    \centering
    \begin{tikzpicture}
        \begin{groupplot}[
            width=.5\linewidth,
            height=.3\linewidth,
            grid=both,
            ymin=5000,ymax=25000,
            xlabel={VSR length $w$},
            group style={
                group size=2 by 1,
                horizontal sep=3mm,
                xticklabels at=edge bottom,
                yticklabels at=edge left
            }
        ]
            \nextgroupplot[
                ylabel={SVSPS},
                legend columns=2,
                legend entries={E+C,E+I+C,All,E+I+X},
                legend pos=south west
            ]
            \addplot table [x=nVoxels,y=overallVoxelStepsPerSecond,dne={builder.springScaffoldings}{EC}] {data/control-scaff.txt};
            \addplot table [x=nVoxels,y=overallVoxelStepsPerSecond,dne={builder.springScaffoldings}{EIC}] {data/control-scaff.txt};
            \addplot table [x=nVoxels,y=overallVoxelStepsPerSecond,dne={builder.springScaffoldings}{EIXC}] {data/control-scaff.txt};
            \addplot table [x=nVoxels,y=overallVoxelStepsPerSecond,dne={builder.springScaffoldings}{EIX}] {data/control-scaff.txt};
            \nextgroupplot[
                legend columns=2,
                legend entries={\SI{0.05}{\second},\SI{0.1}{\second},\SI{0.15}{\second},\SI{0.2}{\second},\SI{0.25}{\second}},
                legend pos=south west
            ]
            \addplot table [x=nVoxels,y=overallVoxelStepsPerSecond,dne={settings.stepFrequency}{0.005}] {data/control-step.txt};
            \addplot table [x=nVoxels,y=overallVoxelStepsPerSecond,dne={settings.stepFrequency}{0.01}] {data/control-step.txt};
            \addplot table [x=nVoxels,y=overallVoxelStepsPerSecond,dne={settings.stepFrequency}{0.015}] {data/control-step.txt};
            \addplot table [x=nVoxels,y=overallVoxelStepsPerSecond,dne={settings.stepFrequency}{0.02}] {data/control-step.txt};
            \addplot table [x=nVoxels,y=overallVoxelStepsPerSecond,dne={settings.stepFrequency}{0.025}] {data/control-step.txt};
        \end{groupplot}
    \end{tikzpicture}
    \caption{
        Simulation performance of a worm-shaped VSR of $w \times 3$ voxels performing locomotion: number of simulated voxel steps per second (SVSPS) vs.\ the length $w$.
        On the left, for different scaffolding configurations; on the right, for different values of the time-step $\Delta t$.
    }
    \label{fig:control}
\end{figure*}
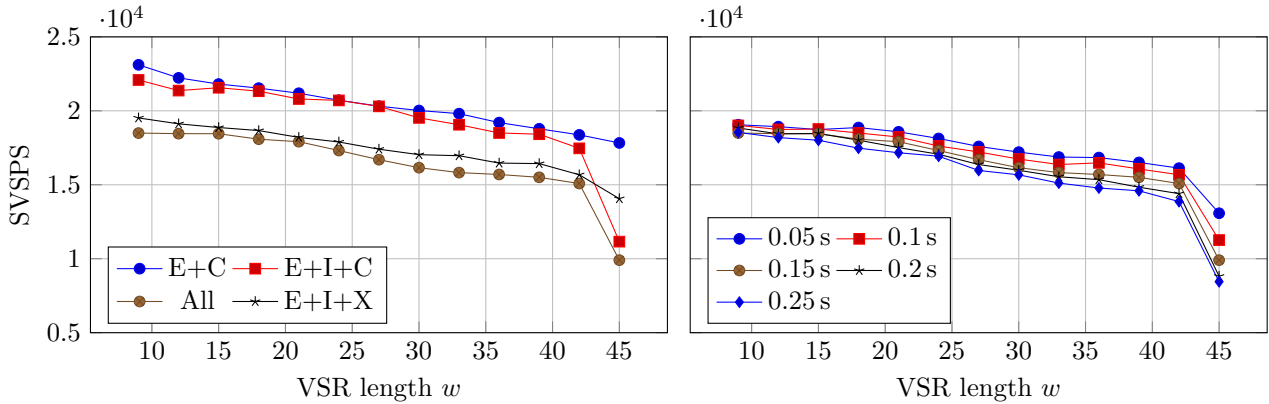

\Cref{fig:control} shows the results in terms of SVSPS vs.\ the VSR length $w$ for different configurations of the scaffolding (left plot) and for different values of the time-step $\Delta t$ (right plot).
It can be seen that \swname{} is able to perform approximately \num{20000} SVSPS per core on the used machine---we remark that each simulation has been executed on a single core.
Moreover, \Cref{fig:control} shows that the number of SVSPS depends on the number of voxels: larger VSRs result in fewer SVSPS.
The remarkably lower values of SVSPS for largest $w$ values in both plots are related to how we executed the experiments: the Java VM took some time to warm up and delivered worse performance in the first executed simulations, that were the ones with $w=45$.

Concerning the impact of the scaffolding, it can be seen that, as expected, the larger the number of simulated SDSs per voxel, the fewer SVSPS: we recall that E+C, E+I+C, All, and E+I+X correspond to $6$, $10$, $18$, and $16$ SDSs, respectively.
Finally, concerning the time-step $\Delta t$, it can be seen that the lower $\Delta t$, the more SVSPS, though the differences are small.
This finding can be explained by considering that the underlying physics engine is required to perform heavier computation when longer time-steps are performed.
We recall, however, that the overall number of performed steps is inversely proportional to $\Delta t$: this makes, with other parameters being the same, convenient to prefer large values for $\Delta t$.

\section{Illustrative examples}
\label{sec:examples}
In order to verify that \swname{} can be a useful tool for research about optimization of VSRs, we repeated the experiments carried out in three significant and recent papers on this topic, namely \cite{hiller2012automatic,kriegman2018morphological,talamini2019evolutionary}.
The cited studies tackled different research questions, but share the fact that they used optimization of VSRs as an experimental way for providing answers.
In~\cite{hiller2012automatic} the investigation is about the possibility of obtaining functioning objects by optimizing the distribution of the materials.
In~\cite{kriegman2018morphological} the question is whether development (i.e., change in the properties of the agent during its lifetime) can increase evolvability.
Finally, in~\cite{talamini2019evolutionary} the hypothesis is that when a VSR controller can sense the environment, it can be more effective.
Interestingly, all the three studies rely on evolutionary computation for doing optimization.

We remark that our intent was not to exactly reproduce the experimental results of the cited studies (also because they were obtained with 3-D VSRs and \swname{} work with 2-D VSRs), but instead to show how \swname{} can be used in a variety of scenarios for a variety of purposes.
To this end, we performed our experiments in similar, but not identical, settings.
In particular, we used here one single evolutionary algorithms (EA) for the three cases and adapted the representation of the solution, and hence the search space, to the specific case.

The used EA employs a $\mu + \lambda$ generational model (see~\cite{de2006evolutionary} for a general introduction to evolutionary computation): a population of $\mu=n_\text{pop}$ individuals is evolved by generating, at each iteration, an offspring of $\lambda=n_\text{pop}$ that is then merged with the parents; the $n_\text{pop}$ best individuals are selected for keeping the population size the same.
The initial population is filled randomly, i.e., by randomly sampling the search space.
The offspring is generated by repeating, for each children, the following steps:
\begin{inparaenum}[(i)]
    \item a genetic operator between crossover and mutation is chosen randomly with probability of, respectively, $0.8$ and $0.2$;
    \item two or one parents, depending on the genetic operator, are chosen using a tournament selection of size $n_\text{tour}$;
    \item the operator is applied and the child is added to the offspring.
\end{inparaenum}
In order to prevent premature convergence to local optima~\cite{squillero2016divergence}, diversity is favored in the population by repeating the process above for at most $10$ times if the individual generated by applying the genetic operator is already present in the population (parents or offspring).
Since we applied this EA on numerical search spaces, i.e., in $\mathbb{R}^p$, we used suitable genetic operators.
For the mutation, we used the Gaussian mutation, i.e., a child $\vec{x}'$ is obtained by adding a zero-mean Gaussian noise independently to each element of the parent $\vec{x} \in \mathbb{R}^p$---we set $\sigma=0.15$.
For the crossover, we used the extended segment crossover, i.e., the child $\vec{x}'$ is obtained as $\vec{x}_1+\vec{\alpha} (\vec{x}_2-\vec{x}_1)$, where $\vec{\alpha} \sim U(-1,2)^p$ is a vector of independent values generated by sampling uniformly the interval $[-1,2]$.

In our experiments, we set $n_\text{pop}=250$, $n_\text{tour}=8$, and we stopped the evolution after $n_\text{gen}=500$ iterations.
We made the code of the experimental machinery using this EA together with \swname{} publicly available at \url{https://github.com/ericmedvet/hmsrevo}.
We set $\Delta t=\SI[quotient-mode=fraction]{1/30}{\second}$ and used the same HPC machines of \Cref{sec:performance-validation}.

\subsection{Evolutionary optimization of the body}
\label{sec:example-body}
Hiller and Lipson~\citep{hiller2012automatic} are the authors of one of the seminal papers about VSRs.
They considered recent progresses in fabrication techniques that enabled robots to be composed of soft materials whose properties vary continuously along the robot.
These advancements brought great freedom in the design of the robots along with, however, the challenge of effectively exploiting this freedom.
For addressing this challenge, the authors used optimization to automatically design, starting from a small set of available materials, the body of a VSR for the task of locomotion; the optimization was performed using a software which simulates the locomotion.
At the end, the results were validated by fabricating a few of the obtained designs and measuring the locomotion performance in reality---the actuation was obtained using air pressure changes, by means of a pneumatic chamber.
We here use \swname{} for reproducing the optimization process using an EA, with a solution representation similar to the one of the cited paper.

One of the key contribution of \cite{hiller2012automatic} is the representation of the solution, i.e., the body of a robot, by using Gaussian mixtures.
We defined our representation based on that idea.

First, we defined four materials by setting different voxel and controller properties, as follows:
\begin{inparaenum}[(i)]
    \item a passive hard material, with $f_s=\SI{25}{\hertz}$, all the scaffolding groups enabled, and a control value $f(t)=0$ (i.e., no actuation);
    \item a passive soft material, with $f_s=\SI{5}{\hertz}$, only the side external and central cross groups enabled, and a control value $f(t)=0$;
    \item an active material, with all the parameters set to default values and a control value $f(t)=\sin(2 \pi t)$;
    \item a counter-phased active material, with all the parameters set to default values and a control value $f(t)=\sin(-2 \pi t)$.
\end{inparaenum}
Then, given an enclosing grid of $w \times h$ voxels, we defined a body to be unequivocally described by $5$ tuples for each material, a tuple being $(\hat{x}_i, \hat{y}_i, \sigma_i, q_i), i \in \{1,\dots,5\}$.

For determining the VSR given its $5 \times 4$ tuples, we proceeded as follows:
\begin{enumerate}
    \item for each element of the grid $(x,y) \in \{1,\dots,w\} \times \{1,\dots,h\}$ and for each material $j$, we computed the sum of the corresponding Gaussian mixtures as:
    \begin{equation*}
        g(x,y,j)=\sum_{i=1}^{i=5} \frac{q_i}{\sigma_i \sqrt{2 \pi}} e^{-\frac{d^2}{\sigma_i^2}}
    \end{equation*}
    where $d=\sqrt{(x-\hat{x}_i)^2+(y-\hat{y}_i)^2}$;
    \item then, for each element of the grid, we set the material to the one with the largest $g(x,y,j)$, if $\max_j g(x,y,j) \ge g_0$, or no material (i.e., no voxel at $x,y$), otherwise---$g_0$ is a predefined threshold that we set to $1$;
    \item finally, we took only the largest connected subset of voxels.
\end{enumerate}
In summary, this representation is an indirect generative representation corresponding to the search space $\mathbb{R}^{5 \times 4 \times 4}=\mathbb{R}^{80}$, that we searched with the EA described in \Cref{sec:examples}.
We generated the initial population by randomly sampling $[0,1]^{80}$.

We experimented with two enclosing grids of $5 \times 5$ and $10 \times 10$ and measured the individual fitness as the relative average velocity in a simulation lasting $\SI{60}{\second}$, i.e., the velocity divided by the largest dimension (width or heigth) of the VSR---we actually used negative velocities for casting the problem as a minimization problem.
We used a flat terrain and performed $10$ repetitions of the execution of the EA for each enclosing grid by varying the random seed.

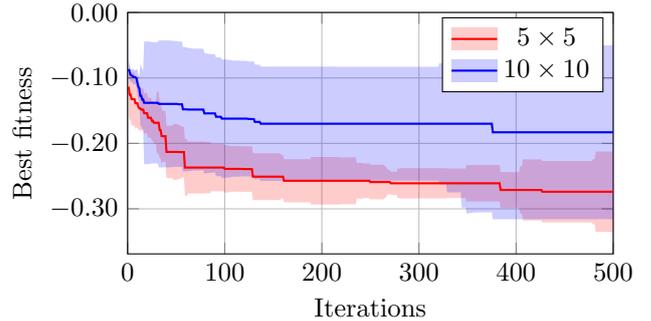
\begin{figure}
    \centering
    \begin{tikzpicture}
        \begin{axis}[
            width=.9\linewidth,
            height=.54\linewidth,
            xlabel=Iterations, ylabel=Best fitness,
            ymax=0,
            xmin=0,xmax=500,
            y tick label style={/pgf/number format/.cd, fixed, fixed zerofill,precision=2,/tikz/.cd},
            grid=both
        ]
            \linewitherrorfilter{data/exp.shape.txt}{iterations}{median}{sd}{shape}{box5x5}{red};
            \linewitherrorfilter{data/exp.shape.txt}{iterations}{median}{sd}{shape}{box10x10}{blue};
            \legend{$5 \times 5$, $10 \times 10$}
        \end{axis}
    \end{tikzpicture}
    \caption{
        Results of the body optimization problem inspired by~\cite{hiller2012automatic}: fitness, i.e., relative average velocity, of the best individual during the evolution.
        The shown value is the median across the $10$ repetitions.
        The shaded area represents the interval defined by $\pm \sigma$, $\sigma$ being the standard deviation across the repetitions.
    }
    \label{fig:exp-shape}
\end{figure}

\Cref{fig:exp-shape} shows the results of this experiments as the median value (across the $10$ repetitions) of the fitness of the best individual, i.e., the one with the largest relative velocity, during the evolution, one line for each of the two enclosing grids.

\begin{figure*}
    \centering
    \includegraphics[width=1\linewidth]{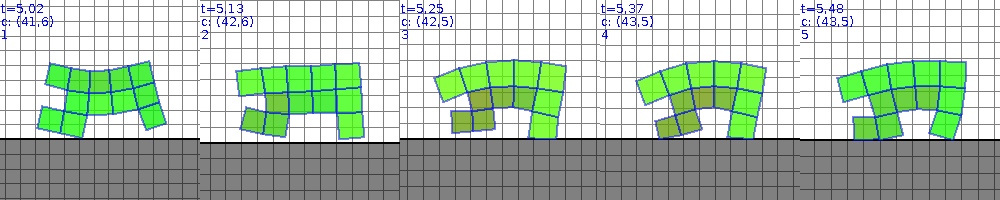}
    \caption{
        Five frames of one of the evolved designs in the $5 \times 5$ case that moves over the flat terrain.
    }
    \label{fig:exp-shape-frames-5x5}
\end{figure*}

It can be seen that the optimization works, i.e., relative velocity is $\approx \times 2.5$ larger in the optimized $5 \times 5$ VSRs and $\approx \times 2$ larger in $5 \times 5$ VSRs with respect to a ``random'' VSR.
Interestingly, the variability of the designs is larger for the largest enclosing grid, which is consistent with the fact that this case allows for more possible shapes.
\Cref{fig:exp-shape-frames-5x5} shows a few frames of one of the evolved designs in the $5 \times 5$ case: the body is actually enclosed in a $5 \times 3$ grid and has two legs, the rear one with a larger ``foot''.
We analyzed the videos and saw that this foot, while moving in coordination with the rest of the body, allows the robot to move fast.

\subsection{Evolution and development}
\label{sec:example-evo-devo}
Kriegman et al.~\cite{kriegman2018morphological} focused on the interaction between evolution and development, two forms of adaptation that occur at different time scales.
For investigating experimentally this topic, the authors considered the case of VSRs: robots had to adapt for locomotion by varying the parameters defining their body and their behavior.

Here we used a very similar representation for the VSRs, all with a body consisting of a fixed grid of $11 \times 4$ voxels with the same properties.
Concerning the controller, we considered two cases: one for the evolution only, the other for combined effect of evolution and development.
In the former case, each $j$-th voxel is actuated by a control value $f_\text{evo}(t)=\sin(2 \pi t + \phi_j)$; in the latter case, the control value is set by $f_\text{evoDevo}(t)=\sin(2 \pi t + \phi_j)+a_j+\frac{t}{t_\text{final}} (b_j-a_j)$, where $t_\text{final}=\SI{60}{\second}$ is the duration of the simulation.
In $f_\text{evoDevo}$ the development is given by the linear variation of the control value, i.e., the relative area of the voxel, that changes ``slowly'' from $a_j$ to $b_j$ over the course of the VSR lifetime.
With this representation a VSR is unequivocally defined by $11 \times 4 = 44$ values $\phi_1,\dots,\phi_{44}$, in the evolution only case, and by $11 \times 4 \times 3 = 132$ values $\phi_1,a_1,b_1,\dots,\phi_{44},a_{44},b_{44}$, in the evolution+development case; the corresponding search spaces are hence $\mathbb{R}^{44}$ and $\mathbb{R}^{132}$.
We generated the initial populations by randomly sampling $[-\pi,\pi]^{44}$ and $[-\pi,\pi]^{132}$.

Similarly to the previous experiment, we used the relative average velocity as the fitness of the individual and we performed $10$ executions of the EA for each of the two cases by varying the random seed.

\begin{figure}
    \centering
    \begin{tikzpicture}
        \begin{axis}[
            width=.9\linewidth,
            height=.54\linewidth,
            xlabel=Iterations, ylabel=Best fitness,
            ymax=0,
            xmin=0,xmax=500,
            y tick label style={/pgf/number format/.cd, fixed, fixed zerofill,precision=2,/tikz/.cd},
            grid=both
        ]
            \linewitherrorfilter{data/exp.evo.devo.txt}{iterations}{median}{sd}{controller}{phasesDevo}{red};
            \linewitherrorfilter{data/exp.evo.devo.txt}{iterations}{median}{sd}{controller}{phases}{blue};
            \legend{Evolution and development, Only evolution}
        \end{axis}
    \end{tikzpicture}
    \caption{
        Results of the evolution vs.\ evolution+development experiment inspired by~\cite{kriegman2018morphological}: fitness, i.e., relative average velocity, of the best individual during the evolution.
        The shown value is the median across the $10$ repetitions.
        The shaded area represents the interval defined by $\pm \sigma$, $\sigma$ being the standard deviation across the repetitions.
    }
    \label{fig:exp-evo-devo}
\end{figure}
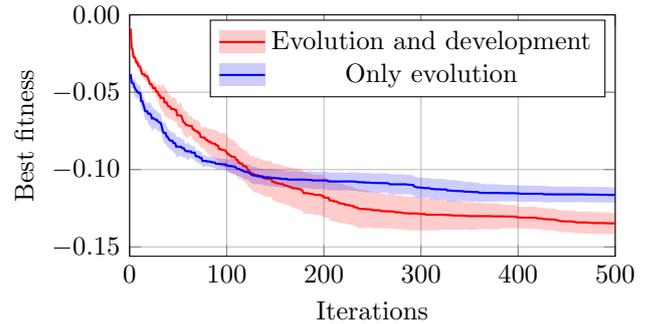

\Cref{fig:exp-evo-devo} shows the results as the median value (across the $10$ repetitions) of the fitness of the best individual during the evolution, one line for each of the two cases.
It can be seen that, in accordance with the findings of the~\cite{kriegman2018morphological}, the combined effect of evolution and development allows to eventually obtain VSRs that are more effective at locomotion.
Another interesting observation concerns the efficiency of the optimization: initially, optimizing in the evolution+development case is harder, likely because the search space is larger.
Qualitatively, this finding is consistent with what Kriegman et al.\ shown experimentally in their study.

\subsection{Evolutionary optimization of sensing controllers}
\label{sec:example-mlp}
Talamini et al.~\cite{talamini2019evolutionary} proposed to design a VSR controller that can exploit the feedback from the environment---that is, sense it---in contrast with existing approaches (as, e.g.,~\cite{hiller2012automatic,kriegman2018morphological}) where the control values were simple functions of the current time.
In order to verify if the ability of sensing actually allows to obtain more effective VSRs, the authors of~\cite{talamini2019evolutionary} considered the locomotion problem, three VSR shapes, and optimized the controller parameters using an EA.
They adopted the stateless, non-sensing controller of~\cite{kriegman2018morphological} (in the evolution only case) and its representation as a comparison baseline.

We here considered three similar shapes:
\begin{inparaenum}[(a)]
    \item a worm of $4 \times 1$ voxels;
    \item a biped with $4 \times 1$ voxels as trunk and two single-voxel legs at the extremes; and 
    \item a tripod with a $5 \times 1$ voxels as trunk and three single-voxel legs, two at the extremes and one in the middle.
\end{inparaenum}
For the baseline controller, the representation is the same of the previous experiment and the search spaces are $\mathbb{R}^4$, $\mathbb{R}^6$, and $\mathbb{R}^8$, respectively for the worm, biped, and tripod.
For the sensing controller, we used a MLP-based controller described in \Cref{sec:vsr-controller} with no inner layer and the following $m=6$ inputs for each one of the VSR voxels:
\begin{inparaenum}[(1)]
    \item the average of the last $5$ readings of the area ratio,
    \item the $5$-th difference of the area ratio,
    \item the average of the last $5$ readings of the rotated $x$-velocity,
    \item the $5$-th difference of the rotated $x$-velocity,
    \item the average of the last $5$ readings of the rotated $y$-velocity, and
    \item the $5$-th difference of the rotated $y$-velocity.
\end{inparaenum}
We set the driving function to $\sin( 2 \pi t)$; this resulted in the MLP being defined by $(6n+1+1)\times n$ weights, $n$ being the number of voxels.
The corresponding search spaces are hence $\mathbb{R}^{104}$, $\mathbb{R}^{228}$, and $\mathbb{R}^{400}$ for the three shapes.
We generated the initial populations by randomly sampling $[-1,1]^{104}$, $[-1,1]^{228}$ and $[-1,1]^{400}$.

We used the relative average velocity as the fitness of the individuals and we performed $10$ executions of the EA for each pair composed of a controller type (sensing or non-sensing) and a shape (worm, biped, or tripod).

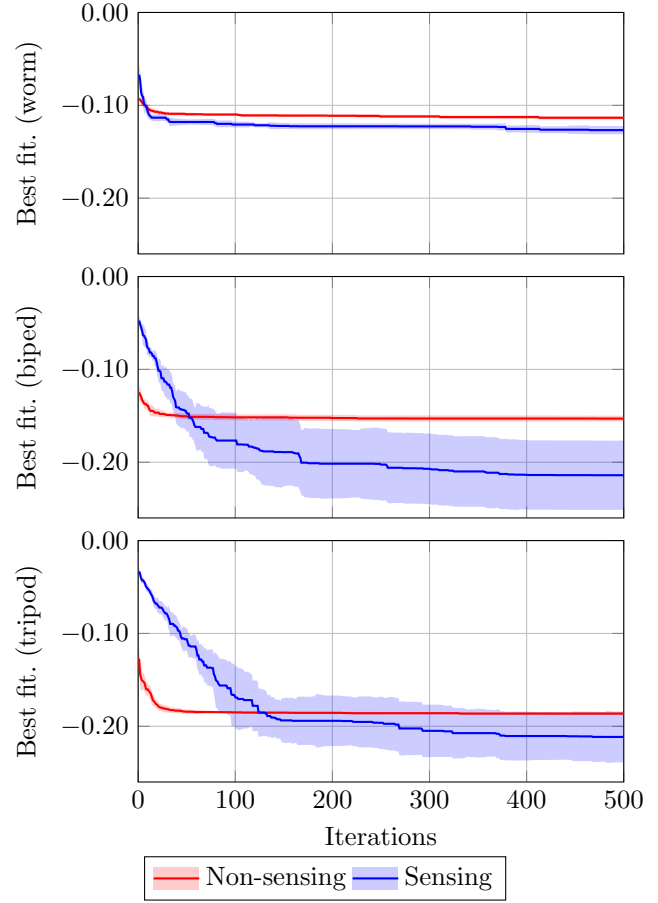
\begin{figure}[!h]
    \centering
    \begin{tikzpicture}
        \begin{groupplot}[
            width=.9\linewidth,
            height=.54\linewidth,
            grid=both,
            ymin=-0.26,ymax=0,
            xmin=0,xmax=500,
            y tick label style={/pgf/number format/.cd, fixed, fixed zerofill,precision=2,/tikz/.cd},
            group style={
                group size=1 by 3,
                horizontal sep=3mm,
                vertical sep=3mm,
                xticklabels at=edge bottom,
                yticklabels at=edge left
            }
        ]
            \nextgroupplot[ylabel={Best fit.\ (worm)},legend columns=2,legend entries={Non-sensing, Sensing},legend to name=outerLegendFitness]
            \linewitherrorfilter{data/exp.controller.worm.txt}{iterations}{median}{sd}{controller}{phases}{red};
            \linewitherrorfilter{data/exp.controller.worm.txt}{iterations}{median}{sd}{controller}{centralizedMLP-0-xya.05md.full}{blue};
            \nextgroupplot[ylabel={Best fit.\ (biped)}]
            \linewitherrorfilter{data/exp.controller.biped.txt}{iterations}{median}{sd}{controller}{phases}{red};
            \linewitherrorfilter{data/exp.controller.biped.txt}{iterations}{median}{sd}{controller}{centralizedMLP-0-xya.05md.full}{blue};
            \nextgroupplot[ylabel={Best fit.\ (tripod)},xlabel=Iterations]
            \linewitherrorfilter{data/exp.controller.tripod.txt}{iterations}{median}{sd}{controller}{phases}{red};
            \linewitherrorfilter{data/exp.controller.tripod.txt}{iterations}{median}{sd}{controller}{centralizedMLP-0-xya.05md.full}{blue};
        \end{groupplot}
    \end{tikzpicture}
    \\
    \ref{outerLegendFitness}
    \caption{
        Results of the sensing vs.\ non-sensing controller experiment inspired by~\cite{talamini2019evolutionary}: fitness, i.e., relative average velocity, of the best individual during the evolution, one plot for each VSR shape.
        The shown value is the median across the $10$ repetitions.
        The shaded area represents the interval defined by $\pm \sigma$, $\sigma$ being the standard deviation across the repetitions.
    }
    \label{fig:exp-mlp}
\end{figure}

\Cref{fig:exp-mlp} shows the results as the median value (across the $10$ repetitions) of the fitness of the best individual during the evolution, one line for each of the two controller cases, one plot for each shape.
It can be seen that the sensing controller is always more effective, after some optimization effort, in accordance with the findings of~\cite{talamini2019evolutionary}.
There are some differences in the amount of improvement that sensing delivers among the three shapes which could be explained, at least in part, by the different increases of the size of the search space.

We visually inspected the behaviors of some of the optimized VSRs and verified that, as found in the cited paper by systematically analyze the behaviors, sensing apparently produces more interesting behaviors.
For example, we discovered that for the worm shape some sensing controllers resulted in a sort of jumping behavior.
The MLP-based controller was able to perceive the change in the value of the $5$-the difference of the rotated $y$-velocity (i.e., acceleration along the $y$-axis) in order to trigger a prompt expansion of all voxels which itself made the VSR jump.
In contrast, many of the non-sensing controllers resulted in a slightly less effective worm-like movement, where the four voxels expand with different phases that cause a sort of wave along the worm body.

\section{Conclusions}
\label{sec:conclusions}
We presented \swname{}, a software for simulating a particular kind of soft robots, called VSRs, composed of many simple voxels.
Our software allows researchers to focus more on what to optimize rather than on how to model a VSR.
In particular, \swname{} may boost research concerning how VSRs can exploit perception and modularity to improve their effectiveness.

We showed that \swname{} is highly versatile and can be easily tailored to a variety of experimental scenarios.
To this end, we used \swname{} for repeating the experiments of three significant and recent studies on VSRs.
The studies are similar, since they all use VSRs and optimization to answer broad research questions, but differ in the nature of the questions they address.

We think that many other interesting research questions still exist that can be tackled also experimenting with VSRs and optimization.
E.g., which kind of development is beneficial to evolution of highly modular robots?
To which degree different kinds of perception may enable the robot to perform more complex tasks?
Can optimization be favored by gradual hardening of the target task? 
Can the body and the brain be optimized together?
Which learning representation for the robot brain result in good scalability with respect to the body size?
We think that, by reducing the burden for the experiments needed to validate answers to these and other questions, \swname{} may become a valuable tool for researchers in different disciplines.

\section*{Conflict of interest}
We wish to confirm that there are no known conflicts of interest associated with this publication and there has been no significant financial support for this work that could have influenced its outcome.

\section*{Acknowledgement}
We thank Giulio Fidel for his contribution to the testing and validation of \swname{}.
The experimental evaluation of this work has been done on CINECA HPC cluster within the CINECA-University of Trieste agreement.

\section*{Current code version}
\label{sec:code-data}
\begin{tabular}{l p{4cm} p{2.5cm}}
    \toprule
    Nr. & Code metadata description & Please fill in this column \\
    \midrule
    C1 & Current code version & v1.0.0 \\
    C2 & Permanent link to repository used for this code version & \url{https://github.com/ericmedvet/2dhmsr/tree/release-1.0.0} \\
    C4 & Legal Code License & GNU GPL v3 \\
    C5 & Code versioning system used & git \\
    C6 & Software code languages, tools, and services used & Java \\
    C7 & Compilation requirements, operating environments \& dependencies & JDK 11, dyn4j \\
    C9 & Support email for questions & \url{emedvet@units.it} \\
    \bottomrule
\end{tabular}

\bibliographystyle{elsarticle-num}
\bibliography{bib}

\end{document}